\documentclass[11pt,a4paper]{article}
\usepackage[hyperref]{acl2020}
\usepackage{times}
\usepackage{latexsym}
\usepackage{amsmath}
\usepackage{amsfonts}
\usepackage{mathrsfs}
\usepackage{amssymb}
\usepackage{url}
\usepackage{pgfplots}
\usepackage{subfigure}
\usepackage{multirow}
\usepackage{latexsym}
\usepackage{booktabs}
\usepackage{algorithm}
\usepackage{algorithmic}

\usepackage{microtype}

\aclfinalcopy 

\title{Knowledge Distillation for \\Multilingual Unsupervised Neural  Machine Translation}

\author{ Haipeng Sun{$^1$}\thanks{\;\;Haipeng Sun was an internship research fellow at NICT when conducting this work.}, Rui Wang{$^2$},  Kehai Chen{$^2$}, \\ \textbf{Masao Utiyama{$^2$}, Eiichiro Sumita{$^2$}, and Tiejun Zhao{$^1$}} \\
	$^1$Harbin Institute of Technology, Harbin, China \\
	$^2$National Institute of Information and Communications Technology (NICT), Kyoto, Japan \\
	\texttt{hpsun@hit-mtlab.net}, \texttt{tjzhao@hit.edu.cn} \\
	\texttt{\{wangrui, khchen, mutiyama, eiichiro.sumita\}@nict.go.jp} \\
	\\}

\date{}

\begin{document}
\maketitle
\begin{abstract}
Unsupervised neural machine translation (UNMT) has recently achieved remarkable results for several language pairs. However, it can only translate between a single language pair and cannot produce translation results for multiple language pairs at the same time. That is, research on  multilingual UNMT  has  been limited. In this paper, we empirically introduce a simple method to translate between thirteen languages using a single encoder and a single decoder, making use of multilingual data to improve UNMT for all language pairs. On the basis of the empirical findings, we propose two knowledge distillation methods to further enhance multilingual UNMT performance. Our experiments on a dataset with English translated to and from  twelve other languages (including three language families and six language branches) show remarkable results, surpassing strong unsupervised individual baselines while achieving promising performance between non-English language pairs in zero-shot translation scenarios and alleviating poor performance in low-resource language pairs.  
\end{abstract}

\section{Introduction}
Recently, neural machine translation (NMT) has been adapted to the unsupervised scenario in which NMT is trained without any bilingual data. Unsupervised NMT (UNMT)  \cite{DBLP:journals/corr/abs-1710-11041,lample2017unsupervised} requires  only monolingual corpora. UNMT achieves remarkable results by using a combination of diverse mechanisms \cite{lample2018phrase} such as an initialization with bilingual word embeddings, denoising auto-encoder \cite{DBLP:journals/jmlr/VincentLLBM10}, back-translation \cite{P16-1009},  and shared latent representation.  More recently, \citet{DBLP:journals/corr/abs-1901-07291} achieves better UNMT performance by introducing the pretrained language model. However, conventional UNMT can only translate between a single language pair and cannot produce translation results for multiple language pairs at the same time \cite{nlp-2020}. 

Multilingual UNMT (MUNMT) translating multiple languages at the same time can save substantial training time and resources. Moreover, the performance of MUNMT in similar languages can promote each other. Research on MUNMT has been limited and there are only a few pioneer studies. For example,  \citet{ijcai2019-739}  and \citet{sen-etal-2019-multilingual}  proposed a multilingual scheme that jointly trains multiple languages with multiple decoders. 
However, the performance of their MUNMT is much worse than our re-implemented individual baselines (shown in Tables \ref{Tab:all-en-non} and \ref{Tab:all-non-en}) and the scale of their study is modest (i.e., 4-5 languages).

In this paper, we empirically introduce an unified framework to translate among thirteen languages (including three language families and six language branches) using a single encoder and single decoder, making use of multilingual data to improve UNMT for all languages. On the basis of these empirical findings, we propose two knowledge distillation methods, i.e.,  self-knowledge distillation and language branch knowledge distillation, to further enhance MUNMT performance.
Our experiments on a dataset with English translated to and from  twelve  other languages  show remarkable results, surpassing strong unsupervised individual baselines.
This paper primarily makes the following contributions:

\begin{itemize}
	\item We propose a unified MUNMT framework to translate between thirteen languages using a single encoder and single decoder. This paper is the first step of multilingual UNMT training on a large scale of European languages.
	\item We propose two knowledge distillation methods for MUNMT and our proposed knowledge distillation methods consider linguistic knowledge in the specific translation task.
	\item Our proposed MUNMT system achieves state-of-the-art  performance on the thirteen languages. It also achieves promising performance in zero-shot translation scenarios and alleviates poor performance in low-resource language pairs.
\end{itemize}

\section{Background of UNMT}
\label{sec:second}

UNMT can be decomposed into four components:  cross-lingual language model pretraining,  denoising  auto-encoder,  back-translation, and shared latent representations. For UNMT,  two monolingual corpora $X^1=\{X_i^1\}$ and $X^2=\{X_i^2\}$ in two languages $L_1$ and $L_2$
are given. $|X^1|$ and $|X^2|$ are the number of sentences in monolingual corpora $\{X_i^1\}$ and $\{X_i^2\}$ respectively.


\subsection{Cross-lingual Language Model Pretraining}
A cross-lingual masked language model, which can encode two monolingual sentences into a shared latent space, is first trained.
The pretrained cross-lingual encoder is then used to initialize the whole UNMT model \cite{DBLP:journals/corr/abs-1901-07291}. 
Compared with previous bilingual embedding pretraining \cite{DBLP:journals/corr/abs-1710-11041,lample2017unsupervised,P18-1005,lample2018phrase,sun-etal-2019-unsupervised}, this pretraining can provide much more cross-lingual information, causing the UNMT model to achieve better performance and faster convergence.

\subsection{Denoising Auto-encoder}

Noise obtained by randomly performing local substitutions and word reorderings \cite{DBLP:journals/jmlr/VincentLLBM10,DBLP:conf/naacl/HillCK16,DBLP:conf/nips/HeXQWYLM16}, is added to the input sentences to improve model learning ability and regularization.  Consequently, the input data are continuously modified and are different at each epoch.
The denoising auto-encoder model objective function can be minimized by encoding a noisy sentence and reconstructing it with the decoder in the same language:

{\footnotesize
\begin{equation}
\begin{aligned}
\mathcal{L}_{D} 
&=\sum_{i=1}^{|X^1|} -log P_{L_1 \to L_1}(X_i^1|C(X_i^1)) \\&+ \sum_{i=1}^{|X^2|} -log P_{L_2 \to L_2}(X_i^2|C(X_i^2)),
\end{aligned}
\end{equation}}%
where $\{C(X_i^1)\}$ and $\{C(X_i^2)\}$ are noisy sentences. $P_{L_1 \to L_1}$ and $P_{L_2 \to L_2}$ denote the reconstruction probability in language  $L_1$ and $L_2$, respectively.          

\subsection{Back-translation}
Back-translation \cite{P16-1009} plays a key role in achieving unsupervised translation that relies only on monolingual corpora in each language. The pseudo-parallel sentence pairs $\{(M^2(X_i^1),X_i^1)\}$ and $\{(M^1(X_i^2),X_i^2)\}$ produced by the model in the previous iteration are used to train the new translation model. Therefore, the back-translation objective function can be optimized by minimizing:

{\footnotesize
\begin{equation}
\begin{aligned}
\mathcal{L}_{B} &= \sum_{i=1}^{|X^1|} -log P_{L_2 \to L_1}(X_i^1|M^2(X_i^1))\\&+\sum_{i=1}^{|X^2|} -log P_{L_1 \to L_2}(X_i^2|M^1(X_i^2)),
\end{aligned}
\end{equation}}%
where $P_{L_1 \to L_2}$ and $P_{L_2 \to L_1}$ denote the translation probability across the two languages.

\subsection{Sharing  Latent  Representations}
Encoders and decoders are (partially) shared between $L_1$ and $L_2$. Therefore, $L_1$ and $L_2$ must use the same vocabulary. The entire training of UNMT needs to consider back-translation between the two languages and their respective denoising processes. In summary, the entire UNMT model can be optimized by minimizing:

{\footnotesize
\begin{equation}
\begin{aligned}
\mathcal{L}_{all} = \mathcal{L}_{D} + \mathcal{L}_{B}.
\end{aligned}
\end{equation}}

\section{Multilingual UNMT (MUNMT)}
\label{sec:third}
\begin{figure}[t]
\centering
\includegraphics[width=0.8\linewidth]{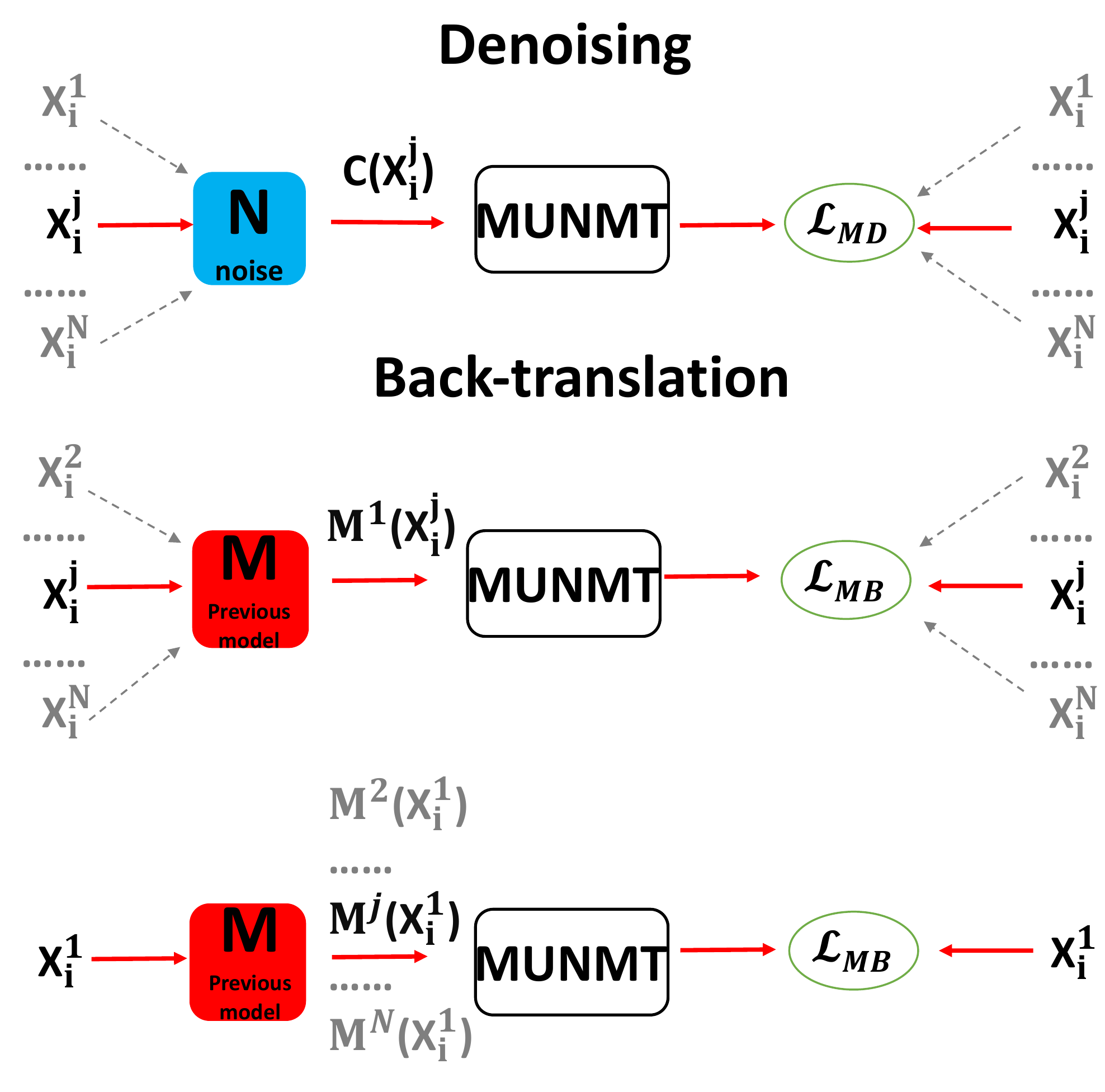}
\caption{\label{fig:MUNMT} {MUNMT architecture.  We take $L_1\leftrightarrow L_j$ time-step as an example. The grey symbols indicate that the  corresponding data are not used or generated during this time-step.}}
\end{figure} 

\subsection{Multilingual Pretraining}
Motivated by \citet{DBLP:journals/corr/abs-1901-07291}, we  construct a multilingual masked language model, using a single encoder. 
For each language, the language model is trained by encoding the masked input and reverting it with this encoder.
This pretrained multilingual language model is used to initialize the full set of parameters of MUNMT.

\begin{figure*}[t]
\centering

\subfigure{
	\begin{minipage}[b]{0.48\linewidth}
	\centering
	\includegraphics[width=0.79\linewidth]{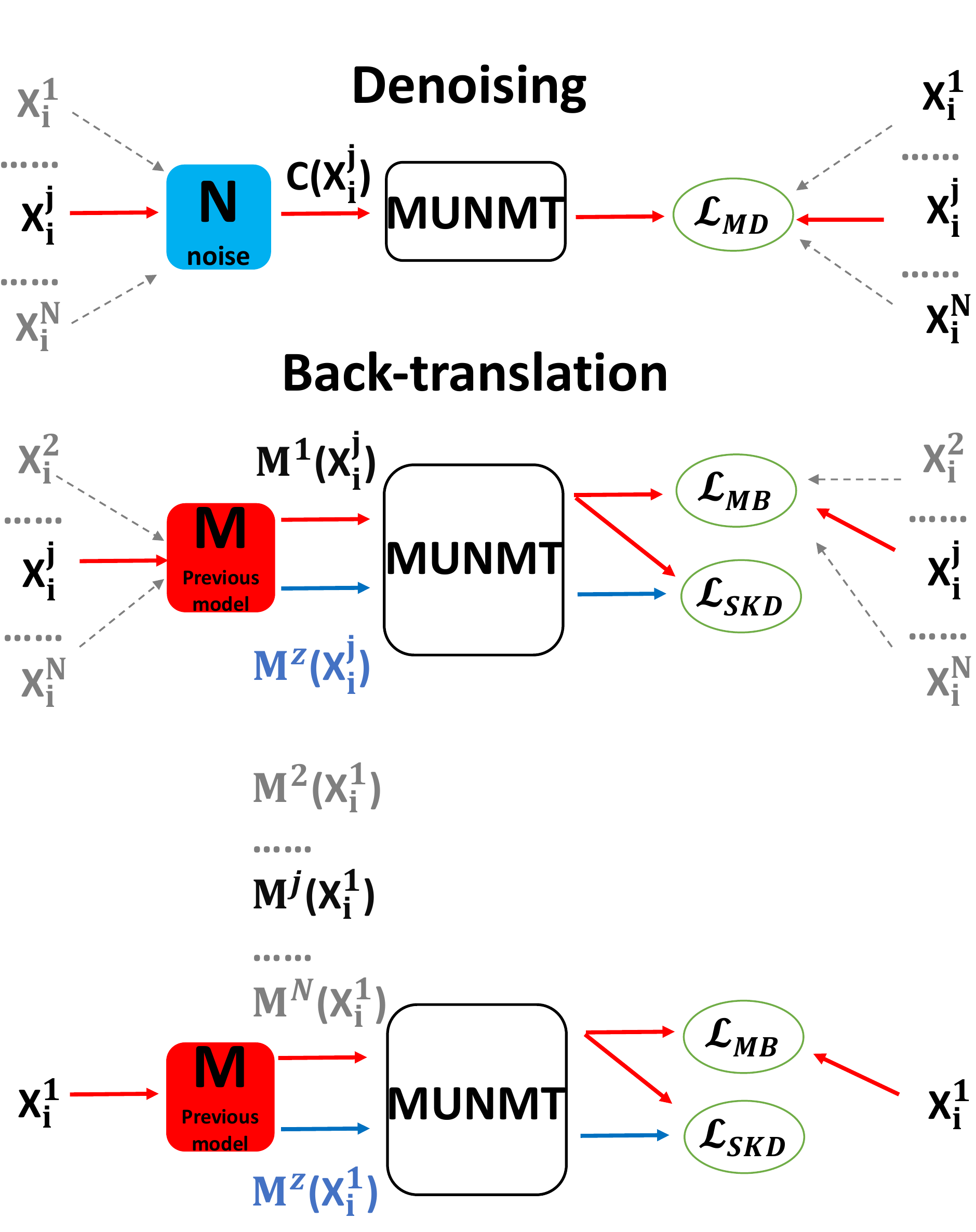}
	\centerline{(a)}\label{fig:SKD}
	\end{minipage}
}
\subfigure{
	
	\begin{minipage}[b]{0.48\linewidth}
	\centering
	\includegraphics[width=0.79\linewidth]{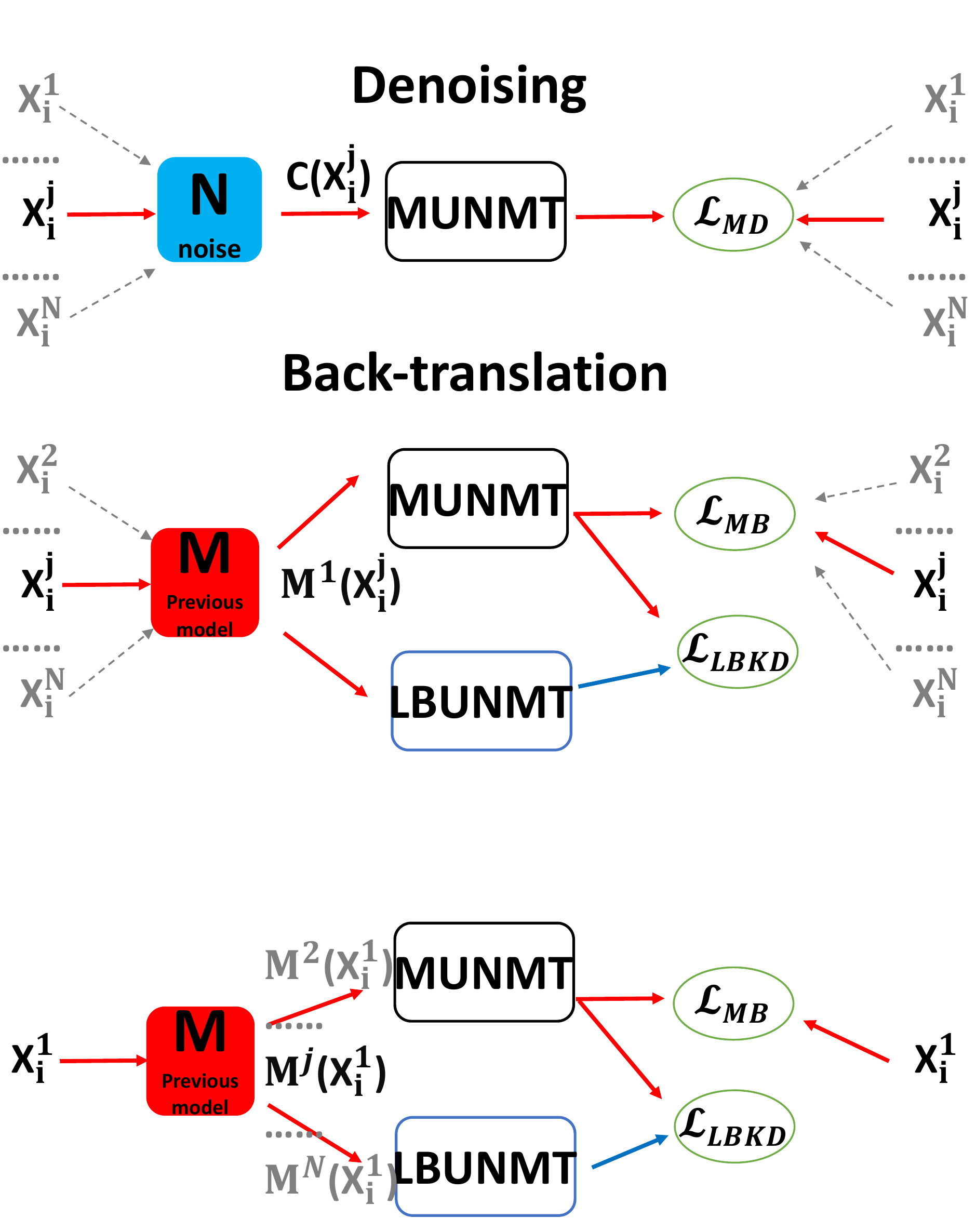}
	\centerline{(b)}\label{fig:LFKD}
	\end{minipage}
}
\caption{(a) Architecture of MUNMT with self-knowledge distillation; (b) Architecture of MUNMT with  language branch knowledge distillation. Similar as Figure \ref{fig:MUNMT}, we take $L_1\leftrightarrow L_j$ time-step as an example. The blue lines denote our proposed knowledge distillation methods are added in the MUNMT training.}
\label{fig:architecture}
\end{figure*}

\subsection{Multilingual UNMT Training}
We have established a MUNMT model on $N$ languages with a single encoder and single decoder.  We denote a sentence  in language $L_j$  as $X_i^j$. For example, $L_1$ indicates English. $|X^j|$ is the number of sentences in the corpus $X^j=\{X_i^j\}$.

As Figure \ref{fig:MUNMT} shows, the entire training process of the MUNMT model is performed  through the denoising and back-translation  mechanisms, between English and non-English language pairs, by minimizing:

{\footnotesize
\begin{equation}
\begin{aligned}
\mathcal{L}_{MUNMT} = \mathcal{L}_{MD} + \mathcal{L}_{MB},
\end{aligned}
\end{equation}}%
where $\mathcal{L}_{MD}$ denotes the denoising function and  $\mathcal{L}_{MB}$ denotes the back-translation function.

In the denoising training, noise (in the form of random token deletion and swapping) is introduced into the input sentences for any language $L_j$.
The denoising auto-encoder, which encodes a noisy version and reconstructs it with the decoder in the same language, is optimized by minimizing:

{\footnotesize
\begin{equation}
\begin{aligned}
\mathcal{L}_{MD} &= \sum_{j=1}^{N} \sum_{i=1}^{|X^j|} -log P_{L_j \to L_j}(X_i^j|C(X_i^j)) ,
\end{aligned}
\end{equation}}%
where $\{C(X_i^j)\}$ is a set of noisy sentences for language $L_j$. $P_{L_j \to L_j}$ denotes the reconstruction probability in  $L_j$. 

In this paper, we primarily focus on the translation from English to other languages or from other languages to English. This is because most test dataset contains English. In the process of back-translation training, we only conduct back-translation from language $L_1$ (English) to other languages and back-translation from other languages to language $L_1$. For any non-English language $L_j$, the pseudo-parallel sentence pairs $\{(M^j(X_i^1),X_i^1)\}$ and $\{(M^1(X_i^j),X_i^j)\}$ are obtained by the previous model in the $L_1 \to L_j$ and  $L_j \to L_1$ direction, respectively. Therefore, the back-translation objective function can be optimized on these pseudo-parallel sentence pairs by minimizing:

{\footnotesize
\begin{equation}
\begin{aligned}
\mathcal{L}_{MB} &=\sum_{j=2}^{N} \sum_{i=1}^{|X^1|} -log P_{L_j \to L_1}(X_i^1|M^j(X_i^1))\\&+\sum_{j=2}^{N}\sum_{i=1}^{|X^j|} -log P_{L_1 \to L_j}(X_i^j|M^1(X_i^j)),
\end{aligned}
\end{equation}}%
where $P_{L_1 \to L_j}$ and $P_{L_j \to L_1}$ denote the translation probabilities, in each direction, between any non-English language and English.

\section{Knowledge Distillation for MUNMT}
\label{sec:fourth}
\begin{algorithm}[tb] 
\caption{The SKD algorithm} 
\label{alg:SKD} 
\begin{algorithmic}[1] 
\REQUIRE ~~\\ 
Monolingual training data $X^1,X^2,\cdots,X^N$;\\
The pretrained model $\theta_0$; 
Number of steps $K$\\
\STATE {Initialize $\theta\leftarrow\theta_0$ }  
\WHILE {Step $q$ $\leq$ max step $K$ }
\FOR{$j=1$; $j<N$; $j++$ }
\STATE Sample batch $\{X_i^j\}$ from $X^j$\\
\STATE Compute denoising loss $\mathcal{L}_{MD}$
\STATE Update $\theta\leftarrow$optimizer($\mathcal{L}_{MD}$)
\ENDFOR
\FOR{$j=2$; $j<N$; $j++$ }
\STATE Sample batch $\{X_i^1\}$from $X^1$\\
\STATE Compute back-translation loss $\mathcal{L}_{MB}$
\STATE Randomly select another language $L_z$ and compute  distillation loss $\mathcal{L}_{SKD}$
\STATE Update $\theta\leftarrow$optimizer($\mathcal{L}_{MB}+\mathcal{L}_{SKD}$)
\STATE Sample batch$\{X_i^j\}$ from $X^j$\\
\STATE Compute back-translation loss $\mathcal{L}_{MB}$
\STATE Randomly select another language $L_z$ and compute  distillation loss $\mathcal{L}_{SKD}$
\STATE Update $\theta\leftarrow$optimizer($\mathcal{L}_{MB}+\mathcal{L}_{SKD}$)
\ENDFOR
		\ENDWHILE  

\end{algorithmic}
\end{algorithm}
To further enhance the performance of  our proposed MUNMT described in Section \ref{sec:third}, we propose two knowledge distillation methods: self-knowledge distillation (Algorithm \ref{alg:SKD}) and language branch knowledge distillation (Algorithm \ref{alg:LBKD}).  
Figure \ref{fig:architecture} illustrates the architecture of MUNMT and the proposed knowledge distillation methods.

Generally, during UNMT training, an objective function $\mathcal{L}_{KD}$ is added, to enhance the generalization ability of the MUNMT model. The general MUNMT objective function can be reformulated as follows:

{\footnotesize
\begin{equation}
\begin{aligned}
\mathcal{L}_{MUNMT} = \mathcal{L}_{MD} + \mathcal{L}_{MB'},\\
\mathcal{L}_{MB'} = (1-\alpha)\mathcal{L}_{MB} + \alpha T^2\mathcal{L}_{KD},
\end{aligned}
\end{equation}}%
where $\alpha$ is a hyper-parameter that adjusts the weight of the two loss functions during back-translation. $T$ denotes the temperature used on the softmax layer. If the temperature is higher,  the  probability distribution obtained would be softer~\cite{DBLP:journals/corr/HintonVD15}.
\subsection{Self-knowledge Distillation}

On the basis of the existing architecture of MUNMT, we introduce self-knowledge distillation \cite{hahn2019self} (SKD) during back-translation, to enhance the generalization ability of the MUNMT model, as shown in Figure \ref{fig:SKD}. Unlike \citet{hahn2019self}'s method, using two soft target probabilities that are based on the word embedding space, we make full use of multilingual information via self-knowledge distillation. 

\begin{figure*}[htbp]
	\centering
	\includegraphics[width=0.9\linewidth]{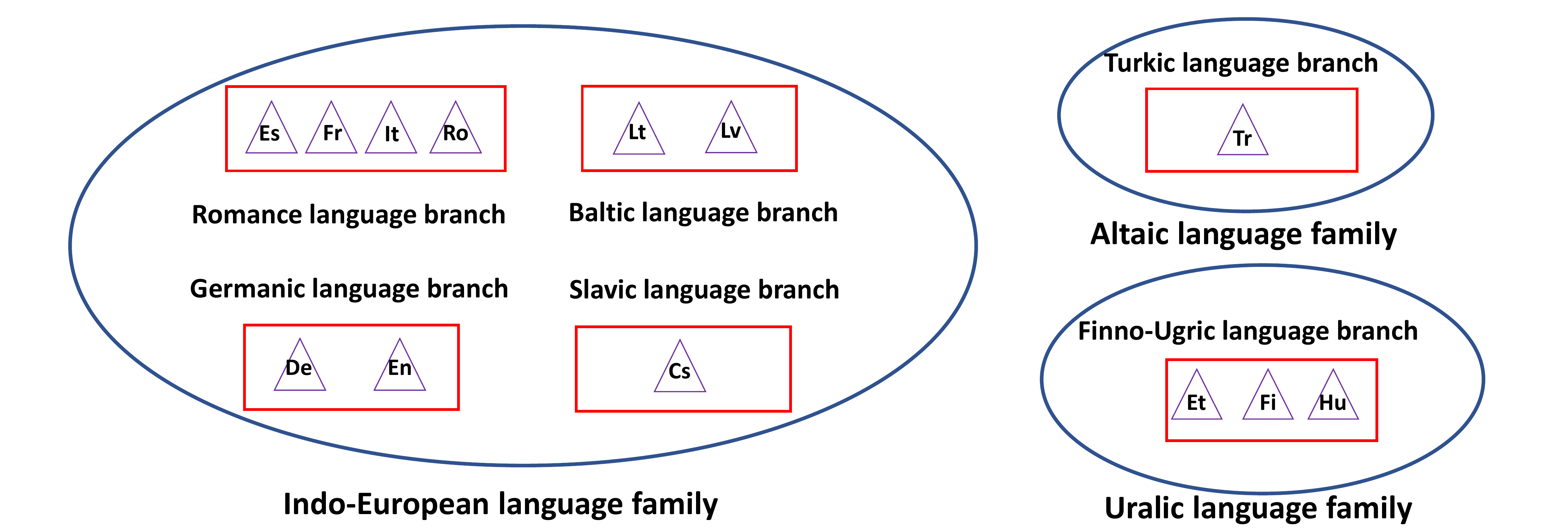}
	\caption{\label{fig:language_family} { The language distribution of our selected languages.}}
\end{figure*}
During back-translation,  only language $L_j$ sentences $M^j(X_i^1)$ are generated before training the MUNMT model in the $L_j \to L_1$ direction. However, other languages, which have substantial multilingual information, are not used during this training. Motivated by this, we propose to introduce  another language $L_z$ (randomly chosen but distinct from $L_1$ and $L_j$) during this training. We argue that the translation from the source sentences through different paths, $L_1 \to L_j \to L_1$ and $L_1 \to L_z \to L_1$, should be similar. The MUNMT model  matches not only the ground-truth output of language $L_j$ sentences $M^j(X_i^1)$, but also the soft probability output of language $L_z$ sentences $M^z(X_i^1)$. The opposite direction is similar.
Therefore, this MUNMT model is optimized by minimizing the objective function:

{\footnotesize
\begin{equation}
\begin{aligned}
&\mathcal{L}_{MB'} = (1-\alpha)\mathcal{L}_{MB} + \alpha T^2\mathcal{L}_{SKD},\\
\mathcal{L}_{SKD} &= \sum_{j=2}^{N}\sum_{i=1}^{|X^1|} KL(X^1(M^j(X_i^1)),X^1(M^z(X_i^1)))                    \\
&+\sum_{j=2}^{N}\sum_{i=1}^{|X^j|} KL(X^j(M^1(X_i^j)),X^j(M^z(X_i^j))),
\end{aligned}
\end{equation}
}%
where $KL(\cdot)$ denotes the KL divergence. It is computed over full output distributions to keep these two probability distributions similar.  For any language $L_j$, $X^1(M^j(X_i^1))$ and $X^1(M^z(X_i^1))$ denote the softened $L_1$ sentence probability distribution after encoding $M^j(X_i^1)$ and $M^z(X_i^1)$, respectively. $M^j(X_i^1)$ and $M^z(X_i^1)$ were generated by the previous model in  the $L_1 \to L_j$ and $L_1 \to L_z$ directions, respectively. $X^j(M^1(X_i^j))$ and $X^j(M^z(X_i^j))$ denote the softened $L_j$  sentence probability distribution after encoding $M^1(X_i^j)$ and $M^z(X_i^j)$, respectively. $M^1(X_i^j)$ and $M^z(X_i^j)$ were generated by the previous model in the $L_j \to L_1$ and $L_j \to L_z$ directions, respectively. Note that zero-shot translation was used to translate language $L_j$ to language $L_z$. The direction $L_j \to L_z$ was not trained during MUNMT training.

\subsection{Language Branch Knowledge Distillation}
 
\begin{algorithm}[tb] 
\caption{ The LBKD algorithm} 
\label{alg:LBKD} 
\begin{algorithmic}[1] 
\REQUIRE ~~\\ 
Monolingual training data $X^1,X^2,\cdots,X^N$;\\
LBUNMT models $\theta_1^{LB},\theta_2^{LB},\cdots,\theta_M^{LB}$;\\
The pretrained model $\theta_0$; 
Number of steps $K$\\
\STATE {Initialize $\theta\leftarrow\theta_0$ }  
\WHILE {Step $q$ $\leq$ max step $K$ }
\FOR{$j=1$; $j<N$; $j++$ }
\STATE Sample batch $\{X_i^j\}$ from $X^j$\\
\STATE Compute denoising loss $\mathcal{L}_{MD}$
\STATE Update $\theta\leftarrow$optimizer($\mathcal{L}_{MD}$)
\ENDFOR
\FOR{$j=2$; $j<N$; $j++$ }
\STATE Sample batch $\{X_i^1\}$from $X^1$\\
\STATE Compute back-translation loss $\mathcal{L}_{MB}$
\STATE Select LBUNMT  language $L_1$ belongs and compute  distillation loss $\mathcal{L}_{LBKD}$
\STATE Update $\theta\leftarrow$optimizer($\mathcal{L}_{MB}+\mathcal{L}_{LBKD}$)
\STATE Sample batch$\{X_i^j\}$ from $X^j$\\
\STATE Compute back-translation loss $\mathcal{L}_{MB}$
\STATE Select LBUNMT language $L_j$ belongs and compute  distillation loss $\mathcal{L}_{LBKD}$
\STATE Update $\theta\leftarrow$optimizer($\mathcal{L}_{MB}+\mathcal{L}_{LBKD}$)
\ENDFOR
		\ENDWHILE  

\end{algorithmic}
\end{algorithm}
We consider thirteen languages: Czech (Cs), German (De), English (En), Spanish (Es), Estonian (Et), Finnish (Fi), French (Fr), Hungarian (Hu),  Lithuanian (Lt), Latvian (Lv), Italian (It), Romanian (Ro), and Turkish (Tr), which belong to three language families including several language branches \cite{lewis2009ethnologue} as shown in Figure \ref{fig:language_family}.

As shown in Figure \ref{fig:LFKD}, we propose knowledge distillation within a language branch (LBKD), to improve MUNMT performance through the existing teacher models. To the best of our knowledge, this is the first proposal that aims to distill knowledge within a language branch. As the number of languages increases, the cost of training time and resources to train an individual model on any two languages increases rapidly.  An alternative knowledge distillation method  within a language branch can avoid this prohibitive computational cost. Because languages  in the same language branch are similar, we first train small multilingual models across all languages in the same language branch (LBUNMT) before training MUNMT. 
The LBUNMT model trained in the same language branch performed better than the single model because similar languages have a positive interaction during the training process as shown in Tables \ref{Tab:all-en-non} and \ref{Tab:all-non-en}.
Therefore, the distilled information of LBUNMT is used to guide the MUNMT model during back-translation. The MUNMT model matches both the ground-truth output and the soft probability output of LBUNMT.
Therefore, this MUNMT model is optimized by minimizing the objective function:

{\footnotesize
\begin{equation}
\begin{aligned}
&\mathcal{L}_{MB'} = (1-\alpha)\mathcal{L}_{MB} + \alpha T^2\mathcal{L}_{LBKD},\\
\mathcal{L}_{LBKD} &=\sum_{j=2}^{N}\sum_{i=1}^{|X^1|} KL(X^1(M^j(X_i^1)),{LB}^1(M^j(X_i^1)))                    \\
&+\sum_{j=2}^{N}\sum_{i=1}^{|X^j|} KL(X^j(M^1(X_i^j)),{LB}^j(M^1(X_i^j))),
\end{aligned}
\end{equation}}%
where $X^1(M^j(X_i^1))$ and ${LB}^1(M^j(X_i^1))$ denote the softened $L_1$ sentence probability distribution of the MUNMT and LBUNMT models, respectively,  after encoding $M^j(X_i^1)$ generated by the previous MUNMT model in the $L_1 \to L_j$ direction. $X^j(M^1(X_i^j))$ and ${LB}^j(M^1(X_i^j))$ denote the softened $L_j$  sentence probability distribution of the MUNMT and LBUNMT models, respectively,  after encoding $M^1(X_i^j)$ generated by the previous MUNMT model in the $L_j \to L_1$ direction.

\section{Experiments}
\label{sec:fifth}

\subsection{Datasets}
To establish an MUNMT system, we considered  13 languages from WMT monolingual news crawl datasets: Cs, De, En, Es, Et, Fi, Fr, Hu, It, Lt, Lv, Ro, and Tr.
For preprocessing, we used the \texttt{Moses} tokenizer \cite{koehn-etal-2007-moses}. For cleaning, we only applied the
\texttt{Moses} script \texttt{clean-corpus-n.perl} to remove lines in the monolingual data containing more than 50 words. We then used a shared vocabulary for all languages, with 80,000 sub-word tokens based on BPE \cite{sennrich2015neural}.
The statistics of the data are  presented in Table \ref{Tab:Statistics}. For Cs,De,En, we randomly extracted 50M monolingual news crawl data after cleaning; For other languages, we used all news crawl data after cleaning as shown in Table \ref{Tab:Statistics}.
\begin{table}[h]
	\center
	\scalebox{.83}{
		\begin{tabular}{l|rrr}
			\toprule
			
			Language & Sentences 	& Words & Sub-words            \\
			\midrule
			Cs & 50.00M 	&  860.36M    &1.16B               \\
			De & 50.00M   &  887.37M    &1.19B                          \\
			En & 50.00M&    1.15B     &1.32B                       \\	 
			Es & 36.33M&    1.01B      &1.19B                          \\
			Et &  3.00M  &  51.39M     &  101.43M                         \\
			Fi & 15.31M   &  189.39M   &  359.78M                         \\
			Fr &  50.00M  &  1.19B   &  1.38B                         \\
			Hu &  34.35M  &  708.13M  &    1.03B                          \\
			It &  30.82M  &  755.56M   &    911.51M                          \\
			Lt &  0.34M  & 6.38M    &   14.64M                          \\
			Lv &  8.60M  &  172.56M    &   281.54M                         \\
			Ro &  8.92M  &    207.07M    &   279.95M                        \\
			Tr &  9.14M  &    153.03M   &    254.70M                       \\

			\bottomrule
	\end{tabular}}
	\caption{Statistics of monolingual corpora.}
	\label{Tab:Statistics}
\end{table}

\begin{table*}[t]
	\center
	\scalebox{.835}{
		\begin{tabular}{l|cccccccc}
			\toprule
			
			Corpus & SNMT&\citet{sen-etal-2019-multilingual} &\citet{ijcai2019-739} & SM & LBUNMT	& MUNMT& SKD  & LBKD   \\
			\midrule
			En-Cs &19.20& -& 6.79  &  14.54 & 14.54	  &  14.40 & 14.89 &     \textbf{15.47}           \\
			En-De & 20.30& 8.09& 13.25 & 18.26& 18.26   &    17.58    & 18.47  & \textbf{19.28}                      \\
			En-Es & 30.40& 14.82 & 20.43& 25.14& 25.40   & 25.05&   25.61  &    \textbf{26.79}                  \\
			En-Et &25.20&  - & -&14.86   &  15.02  & 14.09&   15.03    &   \textbf{15.62}                 \\
			En-Fi & 27.40& -& -  &9.87   &9.99    & 9.75  &   \textbf{10.70}  &     10.57              \\
			En-Fr & 30.60& 13.71& 20.27  &26.02   & 26.36   &25.84    &  26.45  &  \textbf{27.78}                      \\
			En-Hu & -& - & - &11.32   & 11.40   &   10.90   & 11.64   &      \textbf{12.03}      \\
			En-It & -& - & - &24.19    & 24.30   & 23.80     & 24.69  &     \textbf{25.52}         \\
			En-Lt &20.10&  - & - &0.79  & 8.29   &  10.07   &  \textbf{11.15}     &   11.11   \\
			En-Lv & 21.10&- & -   &1.02  & 11.55   & 13.09     & 13.90   &   \textbf{14.33}         \\
			En-Ro & 28.90& -& -  &29.44   &  29.58  & 28.82    & 29.65   &    \textbf{31.28}         \\
			En-Tr & 20.00& - & - &11.87   & 11.87   &  12.41   & 13.24   &    \textbf{13.83}           \\
			\midrule
			Average&  -&- & - &15.61& 17.21& 17.15 & 17.95& \textbf{18.63}\\
			\bottomrule
	\end{tabular}}

	\caption{BLEU scores of all models on the English to non-English language pairs. }
	\begin{flushleft}
		\footnotesize {Note: The first column shows best-performed (till 2019) BLEU scores of supervised NMT (SNMT) systems reported in the corresponding WMT news translation task (\url{http://matrix.statmt.org}).
		The second and third column show BLEU scores reported in the corresponding papers. SM shows the UNMT single model on these two languages (our baseline); LBUNMT shows the multilingual model across all languages in the same language branch; MUNMT shows the multilingual model across all languages; SKD shows the multilingual model with self-knowledge distillation across all languages; LBKD shows the multilingual model with language branch knowledge distillation across all languages. Note that the results for En-Ro are evaluated on the dataset with diacritics removed in the reference text for all our implemented systems.}
	\end{flushleft}
	\label{Tab:all-en-non}
\end{table*}
\begin{table*}[t]
	\center
	\scalebox{.835}{
		\begin{tabular}{l|cccccccc}
			\toprule
			
			Corpus &SNMT& \citet{sen-etal-2019-multilingual} &\citet{ijcai2019-739} &SM & LBUNMT	& MUNMT& SKD& LBKD      \\
			\midrule
			Cs-En & 27.10& - & 11.56 &20.62  & 20.62     &20.09 &21.05 &\textbf{21.25}\\
			De-En & 28.40&11.94 & 16.46  &21.31  &  21.31   &  21.95         & 22.54   &\textbf{22.81}\\ 
			Es-En &31.40&  15.45& 20.35 &25.53& 25.77   &25.37 &26.15 & \textbf{26.59}\\
			Et-En & 30.90&- & -&19.48 &  20.30   & 19.60  &20.95&   \textbf{21.31}\\
			Fi-En &33.00& -& - &7.62  &  7.68    &  7.19  &\textbf{7.92}& 7.80\\
			Fr-En &32.20& 14.47& 19.87 &25.86  &   26.02   & 25.41 & 26.07 & \textbf{26.48}\\
			Hu-En &-& - & - &14.48   &  14.86    & 14.54 & 15.16& \textbf{15.34}\\
			It-En & -&- & - &24.33   &  24.87    &  24.77      & 25.30  & \textbf{25.35}\\
			Lt-En & 36.30& -& -&1.72  & 11.00   & 14.04       &  15.31  & \textbf{15.84}\\
			Lv-En &21.90& - & - &0.95   &   12.75  &  14.90     & \textbf{15.49}   & 15.33\\
			Ro-En &35.20&  - & -&28.52   &   29.57  &  28.38    &  29.58 & \textbf{30.18}\\
			Tr-En & 28.00& - & - &12.99   & 12.99   &  15.65     & 16.85   & \textbf{17.35} \\ 
			\midrule
			Average& -& - & - & 16.95&18.98 & 19.32 & 20.20& \textbf{20.47}\\
			\bottomrule
	\end{tabular}}

	\caption{BLEU scores of all models on the non-English to English language pairs.}
	\label{Tab:all-non-en}
\end{table*}

We report the results for WMT newstest2013 for Cs-En, De-En, Es-En, and Fr-En. We can evaluate the translation performance between pairs of non-English languages because newstest2013 includes these five languages parallel to each other. For other language pairs, we chose the newest WMT newstest set. That is, we reported the results on WMT newstest2019 for Fi-En and Lt-En; WMT newstest2018 for Et-En and Tr-En; WMT newstest2017 for Lv-En; WMT newstest2016 for Ro-En; and WMT newstest2009 for Hu-En and It-En. Note that the versions  of  newstest2019 on Fi/Lt$\rightarrow$ En and En $\rightarrow$ Fi / Lt are different. We chose the corresponding newstest2019 for each direction.

\subsection{Language Model and UNMT Settings}
We used a transformer-based \texttt{XLM} toolkit to train a multilingual masked language model and followed the settings used in \citet{DBLP:journals/corr/abs-1901-07291}: six layers were used for the encoder. The dimension of hidden layers was set to 1024. 
The Adam optimizer \cite{kingma2014adam} was used to optimize the model parameters.
The initial learning rate  was 0.0001, \texttt{$\beta_1 = 0.9$}, and \texttt{$\beta_2 = 0.98$}.

We used the same toolkit and followed the settings of UNMT used in \cite{DBLP:journals/corr/abs-1901-07291}: six layers  were used for the encoder and decoder.  The batch size was set to 2000 tokens. 
The other parameters were the same as those used for training language model. For our proposed knowledge distillation method, $\alpha$ was set to 0.1 and $T$ was set to 2 (the parameters are empirically selected by small-scale experiments and most of the settings achieved good results).
The cross-lingual language model was used to pretrain the encoder and decoder of the whole UNMT model. 
All monolingual data, described in Table  \ref{Tab:Statistics}, were used in the pretraining and MUNMT training phase.
The parameters of the multilingual and single models were the same. 

For evaluation, we used the case-sensitive BLEU scores computed by the \texttt{Moses} script \texttt{multi-bleu.perl}. We executed a single model (two languages) for 60,000 iterations, a small multilingual model (three to five languages) for 30,000 iterations, and a large multilingual model (13 languages) for 15,000 iterations. Eight V100 GPUs were used to train all UNMT models. The single model was trained for approximately two days; the multilingual model (13 languages) costs approximately six days since 13 languages participated in the training.

\subsection{Main Results}

Tables \ref{Tab:all-en-non} and \ref{Tab:all-non-en} present the detailed BLEU scores of all systems on  the English and non-English language pairs, in each direction\footnote{The translation quality of pretrained model was not presented in the Tables \ref{Tab:all-en-non} and \ref{Tab:all-non-en}. The result was poor because the pretrained model (cross-lingual language model) was trained within an encoder. The encoder and decoder of UNMT was initialized with the same parameters of pretrained language model (just an encoder).}.  Our observations are as follows:

1) Our proposed LBUNMT model trained in the same language branch performed better than the single model (SM)  because similar languages have a positive interaction during the training process. Moreover,  SM performed very poorly on  low-resource language pairs such as En-Lt and En-Lv in the Baltic language branch.

2) Our proposed MUNMT model trained in all languages significantly outperformed the previous work \cite{sen-etal-2019-multilingual,ijcai2019-739} by 4$\sim$12 BLEU scores. Moreover, the MUNMT model could alleviate the poor performance achieved with  low-resource language pairs, such as En-Lt and En-Lv. However, the performance of MUNMT is slightly worse than SM in some language pairs.

3) Our proposed knowledge distillation methods outperformed the original MUNMT model by approximately 1 BLEU score. Moreover, our proposed MUNMT with knowledge distillation performed better than SM in all language pairs with fewer training iterations. Regarding our two proposed methods, LBKD achieved better performance since it could obtain much more knowledge distilled from LBUNMT model.

4) There is a gap between the performance of our proposed MUNMT model and that of the supervised NMT systems. To bridge this gap, relying solely on monolingual training data, is worthy of being studied in the future. 

\section{Discussion}
\label{sec:sixth}
\subsection{Zero-shot Translation Analysis}

We also studied the zero-shot translation accuracy of the MUNMT model. Although MUNMT could be trained on all translation directions (ordered language pairs), it would require an extremely long training time. Our proposed MUNMT model was trained in 24 translation directions (all English and non-English language pairs, in each direction), whereas 156 translation directions exist. As the number of languages increases, the number of translation directions increases quadratically.  Therefore, zero-shot translation accuracy is important to the MUNMT model.

\begin{table}[th]
	\centering  \scalebox{.86}{
		\begin{tabular}{lccccc}
			\toprule
			Methods&$\rightarrow$ & Cs	& De & Es & Fr \\
			
			\scalebox{.9}{\citet{ijcai2019-739}} &\multirow{5}{*}{Cs}&-&11.16&11.29&10.61\\
			\scalebox{.9}{\citet{sen-etal-2019-multilingual}}&    &      -         &    -               &     -              &           -                           \\         
			MUNMT&                &  - &     11.91              &        15.22           &  14.66                                    \\
			LBKD&                &  - &      13.16            &       16.63          &                   16.28                   \\   
			SKD&                &  - &      \textbf{16.96}             &         \textbf{20.52}         &             \textbf{20.14}                        \\    
			\midrule
			\scalebox{.9}{\citet{ijcai2019-739} }& \multirow{5}{*}{De}&10.52&-&13.68&9.45\\
			\scalebox{.9}{\citet{sen-etal-2019-multilingual}} &  &      -           &           -        &   7.40                &       6.78                               \\   
			MUNMT&       &10.56            &          -        &   16.15                &      15.85                              \\
			LBKD&                & 11.53  &    -               &     17.27            &                             16.96       \\    
			SKD&                &\textbf{14.58}   &    -               &      \textbf{20.20}            &    \textbf{20.61}                                  \\                         
			\midrule
			\scalebox{.9}{\citet{ijcai2019-739}} &\multirow{5}{*}{Es} &8.32&11.20&-&24.13\\
			\scalebox{.9}{\citet{sen-etal-2019-multilingual}}& &    -              & 4.78                  &                -   &   13.92                                   \\    
			MUNMT&       & 10.04           &      11.87             &        -           &  21.90                                    \\
			LBKD&                & 10.86  &      12.98          &   -               &                                23.05     \\    
			SKD&                & \textbf{13.63}  &        \textbf{16.62}          &   -               &   \textbf{27.04}                                   \\                         
			\midrule
			\scalebox{.9}{\citet{ijcai2019-739}} &\multirow{5}{*}{Fr}&8.89&11.24&23.88&-\\
			\scalebox{.9}{\citet{sen-etal-2019-multilingual}} &  &    -             &    4.59               & 13.87                  &       -                               \\    
			MUNMT&         &   9.77       &           11.70        &     22.30              &    -                                  \\
			LBKD&                & 10.48 &         12.67         &      22.65         & -                                     \\    
			SKD&                & \textbf{13.04}  &        \textbf{16.31}           &           \textbf{25.92}       &                                -      \\                        
			\bottomrule                
	\end{tabular}}
	\caption{BLEU scores of the MUNMT model between pairs of non-English languages. The first two rows of each block are the reported BLEU scores from the corresponding papers. }
	\label{Tab:non-English}
\end{table}

Table \ref{Tab:non-English} shows the performance of translation between non-English language pairs in the zero-shot translation scenario. 
Note that \citet{ijcai2019-739} (2019) shows the results of direct translation between the two languages,  not the result of zero-shot translation.
Compared with previous works, our MUNMT model outperformed  the  previous  systems in almost  all translation directions, particularly the direct translation results reported in \citet{ijcai2019-739}. Compared with the original MUNMT model, our proposed knowledge distillation methods further improved the performance of zero-shot translation.
Regarding   our   two   proposed   methods, SKD significantly outperformed LBKD by  approximately 3  BLEU  scores since the third language was introduced during SKD translation training for two language pairs, achieving much more cross-lingual knowledge. 

\subsection{Further Training (Fine-tuning) Analysis}

To better assess the effectiveness of our proposed MUNMT model, we further trained the MUNMT and LBKD model individually on each language pair for 15,000 iterations. As shown in Tables \ref{Tab:ft-en-non} and  \ref{Tab:ft-non-en}, after further training, the  model outperformed the original single model  on each language pair by approximately 4 BLEU scores. Actually, the number of  iterations of  the whole process (including training the MUNMT model) is half that of the original single model. This demonstrates that our proposed MUNMT model is a robust system and contains substantial cross-lingual information that could improve translation performance. 
\begin{table}[ht]
	\center
	\scalebox{0.85}{
		\begin{tabular}{l|ccccc}
			\toprule
			
			Corpus  & SM 	& MUNMT& +FT   & LBKD& +FT   \\
			\midrule
			En-Cs   &  14.54 	  &  14.40 &    15.79  &  15.47  &  15.93    \\
			En-De  & 18.26   &    17.58    &  19.57  & 19.28  &     20.00              \\
			En-Es & 25.14   & 25.05&    27.59      &  26.79  &  27.80       \\
			En-Et &14.86     & 14.09&   16.62      &  15.62  &    17.21        \\
			En-Fi   &9.87       & 9.75  &  11.05   &  10.57 &     11.58         \\
			En-Fr   &26.02     &25.84    &  28.56   &  27.78&      28.62              \\
			En-Hu  &11.32     &   10.90   &  12.77  &  12.03 & 13.12      \\
			En-It  &24.19       & 23.80     &    25.25 &  25.52 &  25.98      \\
			En-Lt  &0.79    &  10.07   &   10.92     & 11.11 & 11.22 \\
			En-Lv   &1.02    & 13.09     &     14.33 &  14.33 &  15.17     \\
			En-Ro   &29.44    & 28.82    &    32.38  &  31.28 & 32.43  \\
			En-Tr  &11.87    &  12.41   &    14.78  &   13.83  &  15.30      \\
			\midrule
			Average & 15.61 & 17.15 &  19.13 &18.63& 19.53\\
			\bottomrule
	\end{tabular}}

	\caption{The +FT column shows BLEU scores from further training of the MUNMT and LBKD model on the English to non-English language pairs. The other columns show results from Table~\ref{Tab:all-en-non}.   }
	\label{Tab:ft-en-non}
\end{table}

\begin{table}[ht]
	\center
	\scalebox{.85}{
		\begin{tabular}{l|ccccc}
			\toprule
			
			Corpus  & SM	& MUNMT& +FT & LBKD& +FT   \\
			\midrule
			Cs-En  &20.62      &20.09 & 21.50&21.25&22.17\\
			De-En   &21.31    &  21.95& 22.41&22.81&23.07\\
			Es-En  &25.53   &25.37 & 26.24&26.59& 26.78\\
			Et-En &19.48   & 19.60  & 21.61&21.31&22.61\\
			Fi-En  &7.62    &  7.19  & 8.06&7.80& 8.34 \\
			Fr-En  &25.86     & 25.41 & 26.30 &26.48& 26.76\\
			Hu-En  &14.48       & 14.54 &  15.99&15.34&16.07\\
			It-En  &24.33      &  24.77      &  25.54&25.35&25.86 \\
			Lt-En &1.72     & 14.04       & 15.27  &15.84& 16.86 \\
			Lv-En  &0.95     &  14.90     &  15.57 &15.33& 15.87\\
			Ro-En &28.52     &  28.38    &  29.61 &30.18& 30.39\\
			Tr-En &12.99     &  15.65     &  18.47  &17.35& 19.48  \\ 
			\midrule
			Average & 16.95 & 19.32 &  20.55&20.47& 21.19\\
			\bottomrule
	\end{tabular}}
    \caption{The +FT column shows BLEU scores from further training of the MUNMT and LBKD model on the non-English to English language pairs. The other columns show results from Table~\ref{Tab:all-non-en}.}
\label{Tab:ft-non-en}
\end{table}
\section{Related Work}
\label{sec:seventh}
Multilingual NMT  has attracted  much attention  in  the  machine translation community.
\citet{dong-etal-2015-multi} first extended NMT from the translation of a single language pair to multiple language pairs, using a shared encoder and multiple decoders and multiple attention mechanisms, for each language. 
\citet{DBLP:journals/corr/LuongLSVK15}  translated multiple source languages to  multiple target languages using a combination of multiple encoders and multiple decoders.
\citet{firat-etal-2016-multi}  used a shared attention mechanism but multiple encoders and  decoders for each language.
\citet{DBLP:journals/corr/HaNW16}  and \citet{DBLP:journals/tacl/JohnsonSLKWCTVW17}  proposed a simpler method to use one encoder and one decoder to translate between multiple languages. 
Recently, many methods \cite{lakew-etal-2018-comparison,platanios-etal-2018-contextual,sachan-neubig-2018-parameter,blackwood-etal-2018-multilingual,DBLP:conf/wmt/LuKLBZS18,wang19iclr,aharoni-etal-2019-massively,wang-etal-2019-compact,wang-neubig-2019-target} have been proposed to boost multilingual NMT performance. 
In particular, Tan et al.  proposed a knowledge distillation method \citep{DBLP:conf/iclr/TanRHQZL19} and a language clustering method \citep{tan-etal-2019-multilingual} to improve the performance of multilingual NMT. \citet{ren-etal-2018-triangular} propose a triangular architecture to  tackle the problem of low-resource pairs translation by introducing another rich language.

To  further tackle the problem of low-resource pairs translation,
UNMT \cite{DBLP:journals/corr/abs-1710-11041,lample2017unsupervised} has been proposed, using a combination of diverse mechanisms such as initialization with bilingual word embeddings, denoising auto-encoder \cite{DBLP:journals/jmlr/VincentLLBM10}, back-translation \cite{P16-1009},  and shared latent representation.   
\citet{lample2018phrase} concatenated  two bilingual corpora as one monolingual corpus, and used monolingual embedding pretraining in the  initialization step, to achieve remarkable results with some similar language pairs. 
\citet{DBLP:journals/corr/abs-1901-07291}  achieved better UNMT performance by introducing a pretrained language model.
\citet{sun-etal-2019-unsupervised,9043536}  proposed to train UNMT with cross-lingual language representation agreement, to further improve UNMT performance. Moreover, an unsupervised translation task that evaluated in the WMT19 news translation task~\cite{barrault-etal-2019-findings}  attracted many  researchers  to participate~\cite{marie-etal-2019-nicts,li-etal-2019-niutrans}.

For Multilingual UNMT, \citet{ijcai2019-739}  exploited multiple auxiliary languages for jointly boosting UNMT models via the Polygon-Net framework. 
\citet{sen-etal-2019-multilingual}  proposed an MUNMT scheme that jointly trains multiple languages with a shared encoder and multiple decoders. 
In contrast with their use of multiple decoders, we have constructed a simpler MUNMT model with one encoder and one decoder.
Further, we have extended the four or five languages used in their work to thirteen languages, for training our MUNMT model.

\section{Conclusion and Future Work}
\label{sec:eighth}
In this paper, we have introduced a unified framework, using a single encoder and decoder, for MUNMT training on a large scale of European languages.
To further enhance MUNMT performance, we have proposed two knowledge distillation methods. 
Our extensive experiments and analysis demonstrate the effectiveness of our proposed methods. In the future, we intend to extend the work to include language types such as Asian languages.
We will also introduce other effective methods to improve zero-shot translation quality.
\section*{Acknowledgments}
We are grateful to the anonymous reviewers and the area chair for their insightful comments and suggestions.
The corresponding authors are Rui Wang and Tiejun Zhao. Rui Wang was partially supported by JSPS grant-in-aid for early-career scientists (19K20354): ``Unsupervised Neural Machine Translation in Universal Scenarios" and NICT tenure-track researcher startup fund ``Toward Intelligent Machine Translation". Tiejun Zhao was partially supported by National Key Research and Development Program of China via grant 2017YFB1002102. Masao Utiyama was partially supported by  JSPS KAKENHI Grant Number 19H05660.
\bibliography{arxiv}
\bibliographystyle{acl_natbib}

\end{document}